\documentclass[12pt]{article}

\usepackage{arxiv}

\usepackage[utf8]{inputenc} 
\usepackage[T1]{fontenc}    
\usepackage{hyperref}       
\usepackage{url}            
\usepackage{booktabs}       
\usepackage{amsfonts}       
\usepackage{nicefrac}       
\usepackage{microtype}      
\usepackage{lipsum}
\usepackage{multirow}       
\usepackage{graphicx}       
\usepackage{subcaption}     
\usepackage{caption}        

\title{Diagnosing the Impact of AI \\ on Radiology in China}
\author{
    \\
  \textbf{Niklas Muennighoff} \\
  Peking University \\
  muennighoff@stu.pku.edu.cn \\
  Chinese Economy taught by \\
  Li-An Zhou, Hui Wang, Yao Tang \\
}

\begin{document}
\maketitle

\vspace*{25px}

\begin{abstract}

Artificial Intelligence will significantly impact the work environment of radiologists. I suggest that up to 50\% of a radiologists work in 2021 will be performed by AI-models in 2025. However, it won't increase beyond that 50\% level, as radiologists remain key for human-centered aspects of their job. I project that few to no radiologists will be laid off in China due to the existing supply shortage of radiology services in 2021. The application of AI in radiology could contribute 1.7 billion USD to China's GDP in 2025. It will further allow radiologists to start productive work up to four years earlier. AI in radiology will positively impact the health of patients and radiologists themselves.

\end{abstract}

\newpage


\section{Introduction}

In the midst of every crisis, lies great opportunity. The coronavirus pandemic has brought much suffering by killing more than a million people in 2020 \cite{world2020coronavirus}. However, it has accelerated progress in \textbf{medical imaging technology}, such as the analysis of images of lungs \cite{farhat2020deep}. Where exactly does this technology fit in? What are its impacts? 

\subsection{Motivation}

Healthcare can be divided into three parts: Prevention, Diagnosis and Treatment. Diagnosis can be further divided into different methods. Most of these methods, such as MRI-scans\footnote{Magnetic reasonance imaging uses magnets and radio waves to determine body anatomy } or CT scans\footnote{Computed tomography scans use X-rays to locate tissues among other things}, produce image data. This data is then evaluated to reason about medical conditions. It serves as the basis for the third part, the treatment of the patient. The process of dealing with medical scans falls under the discipline of radiology and is carried out by radiologists. Medical imaging technologies aim to make the job of a radiologist easier by facilitating  the creation, evaluation and communication of the image data. Chinese computer scientist Kaifu Lee names the job radiologist as one of the first jobs to be replaced by artificial intelligence \cite{lee2018jobs}. His statement and the recent push in medical technologies forms the motivation of this study. 

\subsection{Why China?}
\label{section:china}

\begin{figure}[h]
    \centering
    \includegraphics[width=14cm]{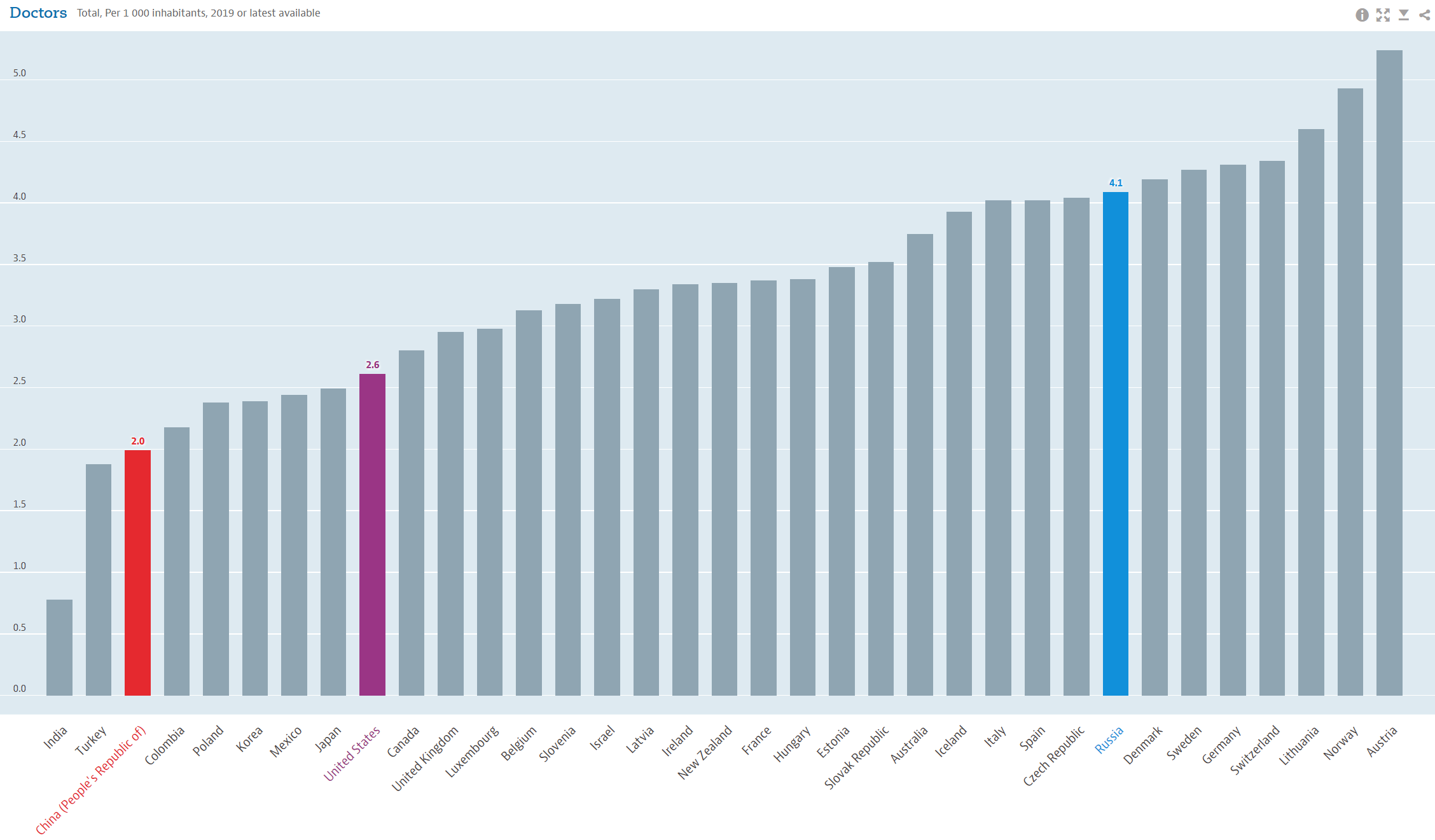}
    \captionsetup{justification=centering}
    \caption{Doctors per capita of selected countries, OECD}
    \label{fig:fig1}
\end{figure}

China is chosen specifically as the subject of this work as I expect the country to be the first to adopt AI-driven radiology for four reasons: 

\begin{itemize}
    \item China has the second largest talent pool of AI Researchers \cite{o2019will}.
    \item China has the largest population in the world and hence the most data. Just feeding more data into AI models has repeatedly shown vast gains in performance \cite{lee2018jobs} \cite{brown2005language}. 
    \item As Figure \ref{fig:fig1} shows, China has less doctors per capita than other developed nations. Hence demand for AI-models to support doctors is much higher than in other countries with comparable developed technologies.
    \item Chinese are comparatively less concerned about AI risks. Research from Deloitte has shown that for many developed nations, such as the US or the UK, a higher percentage of people is concerned about the risks of AI than the percentage of people confident that the country will be able to address AI risks \cite{aisurvey2020deloitte}. Chinese have the lowest concern about AI and are confident in being able to address risks among the countries studied \cite{aisurvey2020deloitte}. Such concerns may put major brakes to AI development in other countries.

\end{itemize}

In the following, I look into scenarios for how artificial intelligence may impact radiology in China. This work aims to highlight both benefits and limits of the technology. I put forward several highlighted recommendations for stakeholders, which are based on my research and understanding of the matter. They should, however, be scrutinized and I am looking forward to critique of my statements.

\section{Paradigm Shift in Medical Imaging}

To be able to assess, we first need to develop an understanding of how artificial intelligence is applied in medical imaging. Subsequently, I will propose scenarios for how it will impact the radiologist job environment, as well as the broader economy and labor markets of China. I will finish the chapter with a study of the potential health impact.

\subsection{Technical Implementation}

\subsubsection{AI-Automation Framework}
\label{section:tech}

\begin{figure}[h]
    \centering
    \includegraphics[width=14cm]{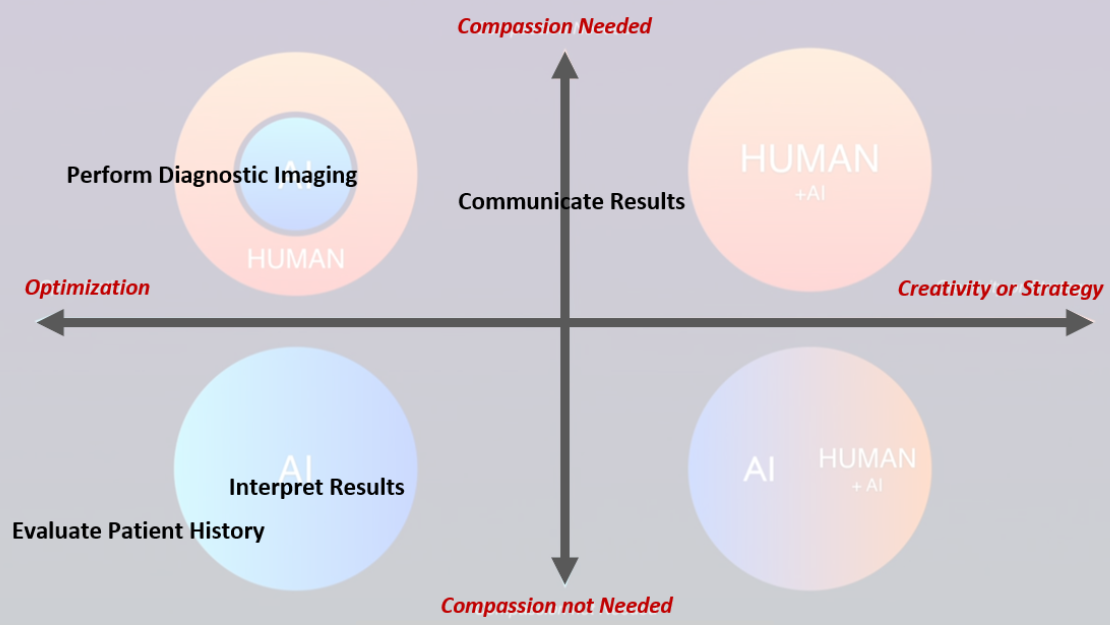}
    \captionsetup{justification=centering}
    \caption{Radiologist tasks grouped into the AI-Automation Framework, Research \& Kaifu Lee}
    \label{fig:fig2}
\end{figure}

The job of a Radiologist can be roughly split into four tasks: Evaluate patient history, perform diagnostic imaging, interpret results, communicate results \cite{wang2013radiology}.
\\
In Figure \ref{fig:fig2}, I classify those tasks into the AI-Automation framework proposed by Kaifu Lee \cite{lee2018jobs}. The x-axis measures how much creativity and strategy is involved in a job. While there has been progress on creative AI-models, such as GANs \footnote{Generative Adversarial Networks were proposed by Ian Goodfellow in 2014} \cite{goodfellow2016nips}, AI-models are best at automating repetitive tasks. The y-axis measures the need for compassion in a job. Especially hospital jobs such as nurse care may need a lot of compassion. Machines have not been shown to be able to express compassion. For each quadrant, Lee assigns an automation scenario, which is superimposed on the graph.
\\  \\
In his own work, Kaifu Lee put the radiologist job into the third quadrant, hence low on creativity and low on compassion. According to the scenario marked in blue, he hence expects the job to be fully replaced by AI. I dig deeper and classify the four identified tasks that make up the radiologist job. The evaluation of a patients history as well as result interpretation are both tasks that exhibit few creativity and do not need a lot of compassion. They can be done alone and are rather analytic, repetitive tasks. Hence they lend themselves to automation. Performing the diagnostic imaging, however, often requires an exchange with the patient. For example, the radiologist may help the patient get into the correct posture before taking a CT-scan. An exchange needs compassion, which according to Lee will remain a human-centered task. I hence put that task in the second quadrant. Similarly, patients may have questions about the results. This also makes a human-to-human interaction necessary for the communication of results. Even if the communication of results happens via email, creativity may be needed to address open-asked questions or format the results in a way the patient can understand (such as for children or the elderly). It may even require strategy, when there are multiple treatment options and the radiologist needs to take the circumstances of the patient into account. Communicating results hence falls between the second and first quadrant and is also difficult to automate.
\\
For further evaluation, I hence make the following two assumptions:
First, I expect two of the four tasks, namely evaluating the patient history and interpreting the results, to be gradually automated. The other two tasks will be fulfilled by humans. Current developments in AI support this classification. As of 2020, state-of-the-art AI models are able to outperform the average radiologist in detecting breast cancer on medical images \cite{rodriguez2019stand}. This falls into the category of interpreting results. Evaluating patient history is a task that may require a combination of text and image data. Recent advancements have closed the gap on text+image understanding tasks between humans and AI \cite{muennighoff2020vilio}. Once a diagnosis for a patient is scheduled, AI models could be queried with the details and then analyze the available data of the patient for medical preconditions that need to be taken into account. Such extraction tasks are often subject of AI research \cite{brown2005language}. The lack of such progress on the other two tasks shows the difficulty of automating them.
\\
Second, I assume that evaluating the patient history and interpreting the results currently make up roughly 50\% of a radiologists job. Hence the average radiologist spends half of his work time on those two tasks. This hypothesis finds some support in literature \cite{gouweloos2019quality}, but is likely to vary a lot between facilities and hence difficult to determine.

\subsubsection{Technology Players}

Taking the technological development into account, I propose the following AI-radiologist workflow: First, an AI-model searches through a patients full history looking for issues that may occur with the planned image taking. Whenever the patient arrives at the clinic, a medical practitioner loads the AI's search results and conducts the diagnostic imaging. The images are fed into the second AI-model, which looks at the images and combined with the history outputs a diagnosis. The practitioner finally communicates the results to the patient.
\\
There are two entry points for technology companies to provide an AI-model. However, the two models need to be able to communicate with each other, as the patient history may also be relevant for the result interpretation done by the second model. Also, the newly interpreted results should be added to the patient history for future evaluation of the first model. While they center around different tasks, Chinese companies have shown their ability to build AI models across different domains. AI platforms from Tencent, Baidu and iFlytek each provide models for more than 50 different tasks as of 2020. This is because different tasks can often be solved by a similar underlying algorithm. One such example is the Attention Algorithm, which is used both for text and image analysis \cite{vaswani2017attention}. Therefore, I expect a single entity to provide models for both tasks to hospitals, rather than two different AI companies specialising on each task. 
\\
One emerging player is Tencent's subsidiary Miying, founded in 2017. At the end of 2020, they reported their algorithms are able to classify 3000 diseases \cite{miying2020news}. Their focus is in line with the two automatable tasks we have identified: First, they are building cloud storage services for patients images and other data. This cloud can then serve as a facility-independent provider of a patient's history. I hypothesize that once they have accumulated significant amounts of patient data, they will provide extraction services leveraging AI search algorithms. Second, they are building AI-models able to detect diseases and hence interpret results of scans.



\subsection{Labor Impact}

\subsubsection{Supply-Demand Projection}

\begin{figure}[h]
    \centering
    \includegraphics[width=14cm]{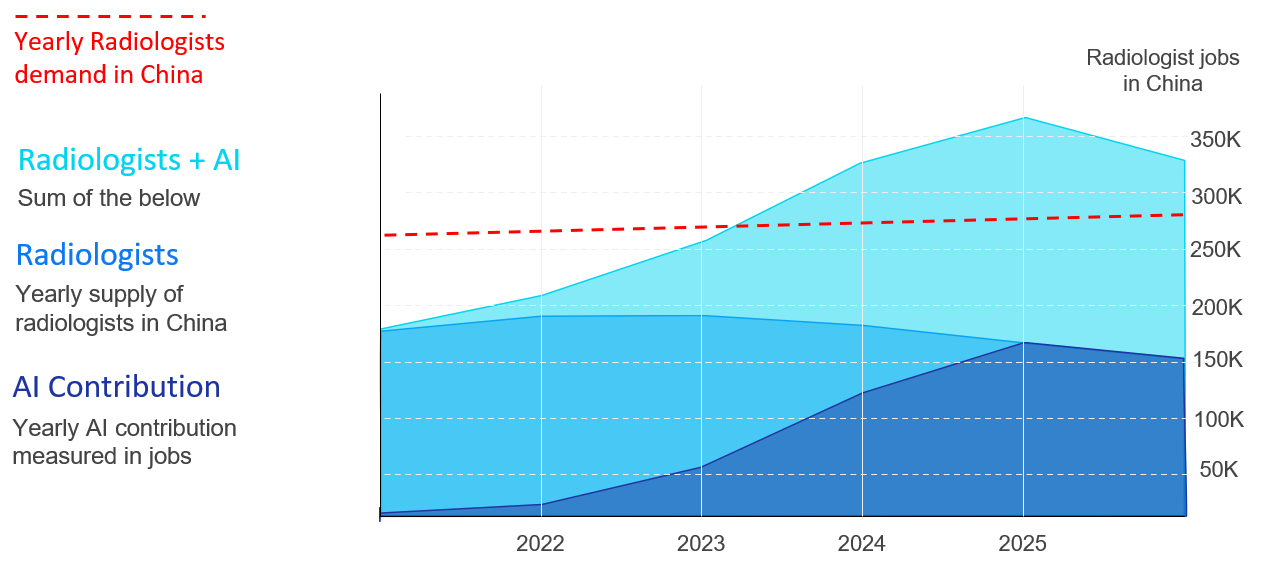}
    \captionsetup{justification=centering}
    \caption{Base-Scenario: Projected Demand/Supply of Radiologists in China, Research}
    \label{fig:fig3}
\end{figure}

In 2017, there were about 160,000 radiologists in China \cite{chinamesa2017radio}. This number has since grown to about 180,000 as of the end of 2020 at an average 4.1\% yearly\footnote{ This estimate ignores the likely spike in temporary radiologists due to the coronavirus pandemic} \cite{china2018data}.
I approach the impact of AI on the radiologist job environment from a demand-supply viewpoint. In Section \ref{section:tech}, I have estimated that AI will be able to take over 50\% of a radiologists job. We can reformulate this as saying that AI will increase the total supply of radiologists by 50\%. This is because once an AI can perform 50\% of a radiologists job, the radiologist can take on 50\% more patients and hence radiologist supply increases. In Figure \ref{fig:fig3}, I graph a possible scenario for the demand \& supply development of the radiologist job in China. \\
2020 and 2019 have been the first two years with notable AI-models that have outperformed the average radiologist \cite{health2019stanford} \cite{rodriguez2019stand}. AI-models can be rolled out very fast, since it is just software that can be executed in the cloud. In 2020, providers of AI-models are in trial stages \cite{daxue2019ai}. Once initial models are successfully rolled out in 2021 and 2022, the scalable nature of software will allow the adoption to accelerate. As Figure \ref{fig:fig3} shows, I therefore expect the 50\% maturity to be reached by 2025. By then AIs will be at the maximum threshold we have established of performing 50\% of a radiologists job. This will effectively double the supply of radiologists.
\\
In 2021, China still has a shortage of radiologists. Figure \ref{fig:fig1} shows that in 2019 China had less than half as many doctors per capita as Russia, and about 20\% less than the United States. Based on that I calculated the shortage for radiologists in China putting the demand between that of the US and Russia. I projected this demand for radiologists to then further increase at China's population growth rate \cite{un2019population}, hence reaching close to 300,000 in 2026. This is shown by the red line in Figure \ref{fig:fig3}. Without AI, the current radiologist growth rate of 4.1\% would close that demand-supply gap in 2030 taking nine years. With AI, however, as Figure \ref{fig:fig3} suggests, China may be able to close that gap within half that time. 

\subsubsection{Labor Market Reactions}

I expect the labor market to react slower than the technological advancement. Initially, this may lead to a surplus of radiologists. However, I do not expect a massive job migration. Rather, fewer radiologists entering the market will naturally decrease supply. This is shown by the change in slope of the human radiologist supply in Figure \ref{fig:fig3} starting in 2023. Due to the current supply shortage, radiologists in China may also be overworking with a report citing a radiologist working 360/365 days per year \cite{chinese2017challenges}. As the supply curve approaches and surpasses the demand, radiologists may adjust their work-life balance leading to a reduction in supply. This will further prevent large-scale layoffs or job migrations. Radiologists who do switch careers will find their existing experience to be of much use in adjacent fields of medicine. There will also remain high demand for them in other less-developed countries, as I expect China to be the pioneer in this field. \\
For radiologists who remain in their job position, their day-to-day tasks will change greatly. They will spend more time with their patients. They will learn how to efficiently interact with the AI-models or even help building them themselves leveraging their knowledge. I \emph{recommend} local governments to support retraining of radiologists, especially in public hospitals where market forces may be slower. I also \emph{recommend} radiologists to educate themselves about how AI can help them or even how they can help AI, as they are a much needed resource to build such models.

\subsubsection{Demand Growth smoothens Disruption}

Among other limitations, projecting demand by population growth in Figure \ref{fig:fig3} may be an understatement for two reasons: 
\begin{enumerate}
\item The affordability of radiology services may increase due to income growth in China. Gross National Income per capita has been growing at about 6\% from 2015 to 2020. If this growth continues and there is no equal growth in prices of radiology, those services will become more affordable and demand may increase \cite{gni2021worldbank}.
\item Demand of radiology services may be much higher for older people. Hence it may correlate more to the growth of the elderly population rather than the growth of the entire population. The United Nations forecasts the share of the 65 or older to increase from 17.4\% in 2020 to 24.8\% in 2030 China \cite{un2019population}.
\end{enumerate}

\begin{figure}[h]
    \centering
    \includegraphics[width=14cm]{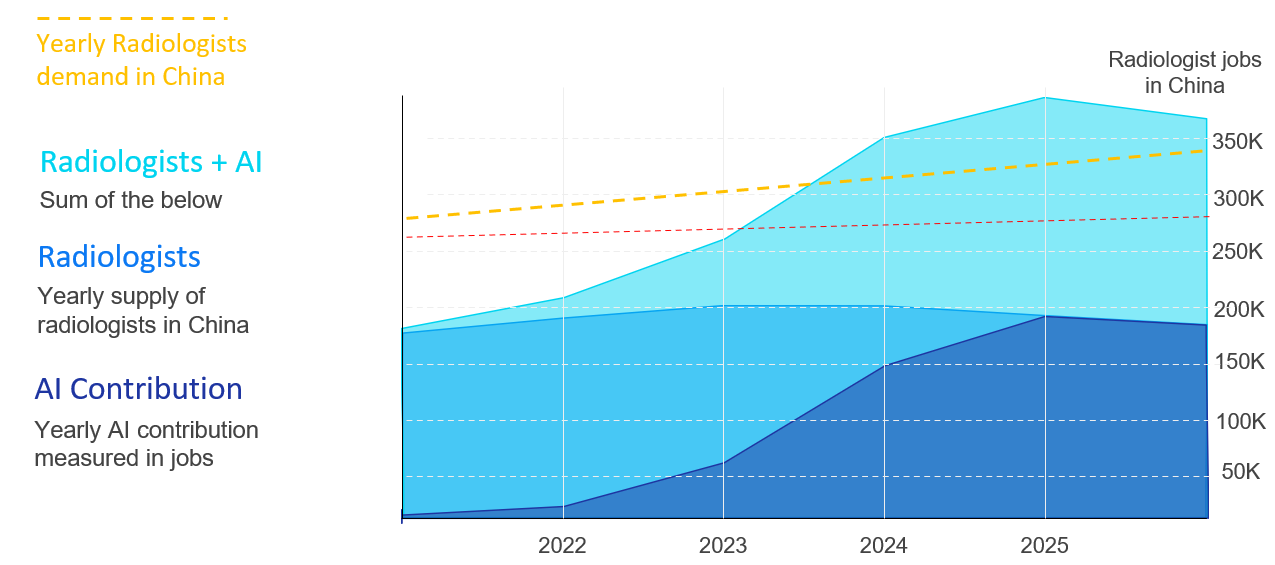}
    \captionsetup{justification=centering}
    \caption{High-Demand Scenario: Projected Demand/Supply of Radiologists in China, Research}
    \label{fig:fig4}
\end{figure}

Based on the latter of those points, Figure \ref{fig:fig4} shows a second scenario with an average 3.79\% growth for radiology services. This is the average yearly growth rate of the population of 65 or older in China between 2020 and 2030. The 3.79\% are about ten times larger than the growth rate for the base scenario, which varies between 0.39\% and 0.15\%. The trend in both scenarios is the same. The higher demand scenario, however, smoothens out the graph. While the AI contribution curve still grows exponential between 2022 and 2024, the higher demand growth rate acts as a buffer for human radiologists. Their decrease in contribution is much slower than in Figure \ref{fig:fig3} and hence fewer of them will have to switch jobs. \\
I infer the following intuitive principle: \textbf{The higher the demand growth rate for a given job, the less disruption will AI cause.} This is because AI will take on some of that additional growth rate, rather than disrupt existing human workers. It is important to note that this is very different from total demand. For total demand, the opposite is true: \textbf{The higher the total demand for a given job, the more disruption will AI cause.}. As AI breaks into sectors with high demand and high employment, the scale of possible layoffs will be much higher and hence cause more disruption. 

\subsection{Economic Impact}

\subsubsection{GDP Contribution of AI in Radiology}
\label{section:gdp}

How could this development affect GDP? The base scenario projects about 170,000 human radiologists in 2025 \footnote{Note that this is less than radiologists in 2020 due to the decrease caused by the AI contribution}. The average yearly salary of a radiologist in China is estimated between 72,000\yen \hspace{0.05cm} and 104,400\yen \hspace{0.05cm} \cite{radiologist2021salary}. We will use 72,000\yen \hspace{0.05cm} as a yearly income estimate. If we assume that a radiologists income roughly equals his GDP contribution, the total 170,000 radiologists will contribute about 1.7 Billion USD in 2025 to China's GDP. In 2020, the average GDP per capita in China was about 70\% higher than the average income per capita, so this assumption is unlikely to hold in practice, but provides a base level estimate \cite{gni2021worldbank}. \\
We have projected that in 2025 AI will provide the other half of the radiologist supply, i.e. there is another 170,000 of AI-driven radiologist supply. If we make a second assumption that the AI GDP-contribution will be equal to that of the human radiologists, AI-driven radiology services will contribute 1.7 Billion USD to China's GDP in 2025. This is about 0.007\% of China's PPP GDP in 2020. \\
PwC estimates the total increase in China's GDP due to AI to be 7.0 trillion USD in 2030 \cite{pwc2021ai}. Given the 1.7 Billion USD, AI-Radiology would make up around 0.027\% of China's AI-driven GDP increase. Given the vague assumptions made, this estimate may vary a lot, but it shows we have been focusing on a tiny part of the bigger revolution that AI is driving. Radiology is just one part of Medical Diagnosis. Medical Diagnosis is also just one part of the Healthcare sector, which is just one of the sectors to be affected by AI. PwC expects the 26\% AI boost to China's GDP to be higher than for the US, Europe or Asia-Pacific. This is in line with the assumption we have made in Section \ref{section:china} that China will be the first place to implement AI-driven radiology.

\subsubsection{Dealing with AI Monopolies}
\label{section:mono}

Where will the new value be created? Just like the Internet, Artificial Intelligence tends to have "Winner takes all" dynamics. Whichever company is able to produce the best model, is likely to get a majority of the customers. This is because the majority of the cost of an AI-model lies in paying for the compute resources to train the model. Once it is trained, the deployment afterwards is either free or requires low-cost cloud services making it very easy to scale. Since the cost of training can reach millions of USD, such as for the GPT-3 model in 2020 \cite{brown2005language}, large companies are at an advantage. Specifically, Tencents subsidiary Miying has access to free cloud services and close to unlimited compute via Tencent Cloud. How stakeholders should deal with such AI monopolies is not straightforward to answer and I leave it for future research to discuss. I can, however, \emph{recommend} governments and other entities to make or even force data to be publicly available. Apart from the compute, an AI-models quality is largely influenced by its input data. The more high-quality data, the better. Making the data publicly available will naturally hinder AI monopolies, as well as foster research as more people can access and build models based on the data. 

\subsubsection{AI impact on Productive Work}

In 2013, Chinese radiologists followed an education program with an average duration of 5 years \cite{wang2013radiology}. Those 5 years have been projected to increase to 7-8 years due to the development of the Chinese Education system and the evolving complexity of medical images. In Section \ref{section:tech} we established that we are able to outsource the complexity of image analysis to AI. Radiologists are left with performing the medical imaging and communicating the results to the patient. This could significantly reduce the time radiologists need to spend in education before starting to work. Let us assume that the 50\% spent on the two tasks that AI will take over, equals to 50\% of their time in education. Learning to correctly interpret a CT-scan for more than 3000 diseases probably needs more learning than standard patient interactions required for the other tasks. However, radiologists will also need to learn new topics relating to the interaction with the AI-models. Hence if we stick with an average 50\% reduction, future radiologists could be educated in half the time, 4 years. This would yield an extra 3-4 years of productive work per radiologist. Additional productive work time is also freed on the patient side. AI-models are likely to be much faster than radiologists in returning results. There is hence less anxious waiting time for patients.

\subsubsection{Embracing AI}
The yearly 4.1\% growth rate of human radiologists compares to a growth rate of medical imaging data of about 30\% in China \cite{china2018data} due to sheer improvements in technology. This simple contrast shows the difficulty for radiologists to manually examine each image. Without automated image processing, radiologists either have to spend less time per image, which will degrade their analysis or they are forced to ignore some images, such as for CT-scans, which can produce more than 700 images per scan. The outcome of both scenarios becomes either a health problem if it worsens radiologists diagnosis or an economic problem if it forces patients to spend time and money to revisit the hospital multiple times until their issue is found. I \emph{recommend} radiologists to embrace AI helping them save lives rather than seeing it as a job threat.

\subsection{Health impacts}

\subsubsection{Access to Radiology Services}

Artificial Intelligence can help increase equal access to radiology services across China. This is because they can be accessed via an internet connection at day and night, which may be required in emergencies. China's internet penetration rate is increasing year-on-year with a recent increase of an absolute 3\% to 64\% in 2020 \cite{statista2020internet}. However, two limits remain: Firstly, AI cannot solve unequal access to imaging machines, such as in provinces like Tibet, which has only 0.94 MRI\footnote{Magnetic Reasonance Imaging} machines per one million people compared to Hainan with 18.66 \cite{chinamesa2017radio}. Secondly, subsidies or special agreements with the providing companies may be necessary to make them available to low-budget hospitals and prevent AI companies from charging monopoly prices.

\subsubsection{Diagnosis Mistakes}

Leveraging AI-models for diagnosis could also reduce diagnosis mistakes. In radiology, those are estimated to make up 3-5\% of all diagnoses \cite{brady2017error}. AI-models in medical imaging are benchmarked against humans and are already able to perform better on classifying certain diseases \cite{rodriguez2019stand}. I go even further and suggest that not only can they help lower diagnosis mistakes, AI-models will help us find new mistakes that we currently classify as a correct diagnosis. While their knowledge may remain narrow in the beginning, future models could incorporate much more cross-disciplinary data and hence point to causes outside of a traditional radiologists knowledge.

\subsubsection{Work-Life Balance of Radiologists}

On the flipside, there may also be health advantages for current radiologists. The Chinese Medical Doctor Association reported in 2017 that radiologists suffer extreme work pressure. They cited a radiologist, who contemplated that he works during night and 360/365 days a year  \cite{chinese2017challenges}. This seems to be in line with the supply-demand gap identified in Figure \ref{fig:fig3}. With the rise of the AI contribution identified in the chart, some of that pressure should be relieved. Further, machines can work at all times, which could allow radiologists to work less during night hours in the future.

\section{Limits \& Discussion}

\subsection{Assumptions}

Throughout this research I have made several assumptions, which should be scrutinized, as they may not hold in practice. We started out with the assumption that two of the radiologists four tasks are not automatable relying on Kaifu Lee's AI-Automation framework in Figure \ref{fig:fig2}. Kaifu Lee himself admits his framework is not perfect \cite{lee2018jobs}. Who can guarantee that machines cannot display compassion? In fact, some humans may even prefer communicating with an AI than a human when it comes to sensible content. This is shown by the more than 2 million people who have downloaded the app Replika, which allows people to talk about their problems with an AI friend. Hence, some patients may prefer the communication of sensible results to be done by an AI, which could raise the 50\% threshold. \\
However, the time horizon for this may be a very different one. The scope of this research focuses on the next 5 - 10 years. In the long-term none of the assumptions made about AI will hold true. This is because if we believe that technological progress will not stop, then it is inevitable that we will eventually be able to build AIs indistinguishable from humans. Whether we will decide to do so, is a different question. \\
The estimates behind the growth rates of Figure \ref{fig:fig3} may vary significantly. We have already looked into variations of the projected demand, but the supply curve may also be different. Dermatologists were the first doctors, whose performance was surpassed by AI in 2018 \cite{dermatologist2021ai}. However, two years later AI has still not penetrated the dermatologist job environment. This is likely due to legacy and reservations about the use of AI. Radiologists have started to be surpassed in 2019 and 2020. Given the prime conditions in China established in Section \ref{section:china}, I have estimated the use of AI in radiology to gain moment around 2023 in Figure \ref{fig:fig3}. Unexpected technical problems, mistrust or government regulations may, however, delay or limit the use of AI.

\subsection{Future Work}

In this work, we have looked at a very tiny field. I propose four directions that are interesting for future work: 

\begin{itemize}
    \item In section \ref{section:gdp}, I estimate radiology to merely make up 0.027\% of the total AI contribution to China's economy. Many other fields may be interesting to research. I suggest: AI in the voice-over industry, AI in logistics or AI in dermatology.
    \item I have mostly focused on one company, Tencent Miying. This is because at this point, AI-Radiology companies are still in stealth mode as they have few if any have daily customers and have not realized any profits yet \cite{daxue2019ai}. Once many hospitals have adopted AI-Radiology services, it may be interesting to look deeper into the competition dynamics. Do companies compete on price or rather model quality? Is a monopoly forming? 
    \item We have tried to quantify as many variables as possible, sometimes requiring vague estimates. However, we have not included the health impact in our analysis of economic impacts. If AI in radiology can lead to faster recovery and heal more people, this could increase the availability of productive workers and hence GDP. Looking into how such health impacts may affect the economy is a topic worth researching.
    \item Will there be AI monopolies and if yes how should we deal with them? I have briefly touched upon this topic in Section \ref{section:mono}. Access to large amounts of computation and data are paramount in AI. This may constitute a barrier of entry for smaller companies.
\end{itemize}

\section{Conclusion}

We started with a technical discussion of the radiology tasks AI can reasonably replace. Based on that, I proposed two scenarios for the supply-demand development of radiology with different demand growth rates. In both scenarios, few workers will be displaced. I established that the higher the growth rate of demand, the lower the disruption caused by AI. The economy will benefit from AI both due to its GDP contribution and freeing up of productive work time. However, dealing with AI monopolies may become an important task. Further, adopting AI in radiology may positively impact the health of current customers, potential customers with limited access and radiologists themselves. \\
I have put forward several recommendations for stakeholders: Radiologists should embrace AI and educate themselves about how it can help them or they can help it. Local governments and other entities should support the retraining of radiologists especially in public hospitals. Further, they should force data to be as accessible as possible and not locked into few companies.

\newpage

\section{Figures}

\newpage


\bibliographystyle{unsrt}  
\bibliography{references} 

\begin{thebibliography}{10}

\bibitem{world2020coronavirus}
World~Health Organization et~al.
\newblock Coronavirus disease (covid-19): weekly epidemiological update.
\newblock 2020.

\bibitem{farhat2020deep}
Hanan Farhat, George~E Sakr, and Rima Kilany.
\newblock Deep learning applications in pulmonary medical imaging: recent
  updates and insights on covid-19.
\newblock {\em Machine vision and applications}, 31(6):1--42, 2020.

\bibitem{lee2018jobs}
Kai-Fu Lee.
\newblock {\em AI superpowers: China, Silicon Valley, and the new world order}.
\newblock Houghton Mifflin Harcourt, 2018.

\bibitem{o2019will}
Sarah O’Meara.
\newblock Will china lead the world in ai by 2030?
\newblock {\em Nature}, 572(7770):427--428, 2019.

\bibitem{brown2005language}
TB~Brown, B~Mann, N~Ryder, M~Subbiah, J~Kaplan, P~Dhariwal, A~Neelakantan,
  P~Shyam, G~Sastry, A~Askell, et~al.
\newblock Language models are few-shot learners. arxiv 2020.
\newblock {\em arXiv preprint arXiv:2005.14165}, 4.

\bibitem{aisurvey2020deloitte}
Deloitte.
\newblock Global ai survey.
\newblock
  \url{https://www2.deloitte.com/content/dam/Deloitte/lu/Documents/public-sector/lu-global-ai-survey.pdf},
  2020.

\bibitem{wang2013radiology}
Ye~Elaine Wang, Mengfei Liu, Long Jin, Matthew~P Lungren, Lars~J Grimm, Ziwei
  Zhang, and Charles~M Maxfield.
\newblock Radiology education in china.
\newblock {\em Journal of the American College of Radiology}, 10(3):213--219,
  2013.

\bibitem{goodfellow2016nips}
Ian Goodfellow.
\newblock Nips 2016 tutorial: Generative adversarial networks.
\newblock {\em arXiv preprint arXiv:1701.00160}, 2016.

\bibitem{rodriguez2019stand}
Alejandro Rodriguez-Ruiz, Kristina L{\aa}ng, Albert Gubern-Merida, Mireille
  Broeders, Gisella Gennaro, Paola Clauser, Thomas~H Helbich, Margarita
  Chevalier, Tao Tan, Thomas Mertelmeier, et~al.
\newblock Stand-alone artificial intelligence for breast cancer detection in
  mammography: comparison with 101 radiologists.
\newblock {\em JNCI: Journal of the National Cancer Institute},
  111(9):916--922, 2019.

\bibitem{muennighoff2020vilio}
Niklas Muennighoff.
\newblock Vilio: State-of-the-art visio-linguistic models applied to hateful
  memes.
\newblock {\em arXiv preprint arXiv:2012.07788}, 2020.

\bibitem{gouweloos2019quality}
Femke~A Gouweloos.
\newblock Quality and efficiency within radiology and the added value of a
  regional pacs.
\newblock Master's thesis, University of Twente, 2019.

\bibitem{vaswani2017attention}
Ashish Vaswani, Noam Shazeer, Niki Parmar, Jakob Uszkoreit, Llion Jones,
  Aidan~N Gomez, {\L}ukasz Kaiser, and Illia Polosukhin.
\newblock Attention is all you need.
\newblock In {\em Advances in neural information processing systems}, pages
  5998--6008, 2017.

\bibitem{miying2020news}
Tencent Miying.
\newblock Official news tencent miying (cn).
\newblock \url{https://miying.qq.com/official/detailnews/654}, 2020.

\bibitem{chinamesa2017radio}
Chinamesa.
\newblock Radiologist distribution data (cn).
\newblock \url{http://www.chinamesa.org/nd.jsp?id=146}, 2017.

\bibitem{china2018data}
CSF-Sim.
\newblock 2018 use of ai technology (cn).
\newblock {\em Life System Modeling and Simulation--China Simulation Society},
  2018.

\bibitem{health2019stanford}
Healthcareitnews.
\newblock Chinese ai beats stanford-team.
\newblock
  \url{https://www.healthcareitnews.com/ai-powered-healthcare/chinese-ai-start-beats-stanford-team-x-ray-diagnostic-competition},
  2019.

\bibitem{daxue2019ai}
Daxue Consulting.
\newblock Ai healthcare china.
\newblock \url{https://daxueconsulting.com/ai-healthcare-china/}, 2019.

\bibitem{un2019population}
United Nations.
\newblock Population forecasts.
\newblock \url{https://population.un.org/wpp/DataQuery/}, 2019.

\bibitem{chinese2017challenges}
Chinese Medical~Doctor Association.
\newblock Challenges faced by radiologists in china (cn).
\newblock \url{http://www.cmda.net/fhzwhdt/11293.jhtml}, 2017.

\bibitem{gni2021worldbank}
Worldbank.
\newblock Gni per capita, china.
\newblock
  \url{https://data.worldbank.org/indicator/NY.GNP.PCAP.KD.ZG?locations=CN},
  2021.

\bibitem{radiologist2021salary}
Jobui 58.com.
\newblock Radiologist salary estimates in china (cn).
\newblock \url{https://www.58.com/zhiwei/fangshekeyisheng/xinzi/},
  \url{https://www.jobui.com/salary/quanguo-fangshekeyisheng/}, 2021.

\bibitem{pwc2021ai}
PWC.
\newblock Ai analysis sizing the prize, china.
\newblock
  \url{https://www.pwc.com/gx/en/issues/analytics/assets/pwc-ai-analysis-sizing-the-prize-report.pdf},
  2020.

\bibitem{statista2020internet}
Statista.
\newblock Internet penetration.
\newblock
  \url{https://www.statista.com/statistics/236963/penetration-rate-of-internet-users-in-china/},
  2020.

\bibitem{brady2017error}
Adrian~P Brady.
\newblock Error and discrepancy in radiology: inevitable or avoidable?
\newblock {\em Insights into imaging}, 8(1):171--182, 2017.

\bibitem{dermatologist2021ai}
USNews.
\newblock Ai beats dermatologists.
\newblock
  \url{https://www.usnews.com/news/health-care-news/articles/2018-05-28/artificial-intelligence-beats-dermatologists-at-diagnosing-skin-cancer},
  2021.

\end{thebibliography}

\end{document}